\documentclass[nohyperref]{article}

\usepackage{xcolor}
\usepackage{microtype}
\usepackage{graphicx}
\usepackage{subfigure}
\usepackage{booktabs} 
\usepackage{listings}

\usepackage{hyperref}

\usepackage[accepted]{icml2022/icml2022}
\makeatletter
\renewcommand{\ICML@appearing}{}
\makeatother

\usepackage{amsmath}
\usepackage{amssymb}
\usepackage{mathtools}
\usepackage{amsthm}

\usepackage[capitalize,noabbrev]{cleveref}

\theoremstyle{plain}
\newtheorem{theorem}{Theorem}[section]

\theoremstyle{definition}
\newtheorem{definition}[theorem]{Definition}

\theoremstyle{remark}

\usepackage[disable,textsize=tiny]{todonotes}

\icmltitlerunning{}

\newcommand{\Aug}{A}
\newcommand{\bigO}{\mathcal{O}}
\newcommand{\calM}{\mathcal{M}}
\newcommand{\clip}{\mathrm{clip}}
\newcommand{\gauss}{\mathcal{N}}
\newcommand{\prob}{\mathbb{P}}
\newcommand{\synlm}{SynLM~}

\newcommand{\tok}{\mathrm{tok}}

\definecolor{codegreen}{rgb}{0,0.6,0}
\definecolor{codegray}{rgb}{0.5,0.5,0.5}
\definecolor{codepurple}{rgb}{0.58,0,0.82}
\definecolor{backcolour}{rgb}{0.95,0.95,0.92}

\lstdefinestyle{mystyle}{
    backgroundcolor=\color{backcolour},   
    commentstyle=\color{codegreen},
    keywordstyle=\color{magenta},
    numberstyle=\tiny\color{codegray},
    stringstyle=\color{codepurple},
    basicstyle=\ttfamily\footnotesize,
    breakatwhitespace=false,         
    breaklines=true,                 
    captionpos=b,                    
    keepspaces=true,                 
    numbers=left,                    
    numbersep=5pt,                  
    showspaces=false,                
    showstringspaces=false,
    showtabs=false,                  
    tabsize=2
}

\begin{document}

\twocolumn[
\icmltitle{Privately generating tabular data using language models}



\icmlsetsymbol{equal}{*}

\begin{icmlauthorlist}
\icmlauthor{Alexandre Sablayrolles}{yyy}
\icmlauthor{Yue Wang}{yyy}
\icmlauthor{Brian Karrer}{yyy}
\end{icmlauthorlist}

\icmlaffiliation{yyy}{Meta AI}

\icmlcorrespondingauthor{Alexandre Sablayrolles}{asablayrolles@meta.com}

\icmlkeywords{private synthetic data generation, language models}

\vskip 0.3in
]



\printAffiliationsAndNotice{}  

\begin{abstract}
Privately generating synthetic data from a table is an important brick of a privacy-first world.  We propose and investigate a simple approach of treating each row in a table as a sentence and training a language model with differential privacy.  We show this approach obtains competitive results in modelling tabular data across multiple datasets, even at small scales that favor alternative methods based on marginal distributions. 
\end{abstract}

\section{Introduction}

Privacy has become a major concern in the collection and usage of data.
In a privacy-first world, the access to raw data is very restricted and limited to infrastructure-related purposes.
For the purpose of training machine learning models or computing analytics, data is accessed through privacy-preserving channels such as differentially-private query engines.
In such an environment, developing new algorithms or debugging code can be challenging, as each query on the data is billed to the privacy budget, and it is impossible to ``eye-ball" data to figure out potential issues and problems.
To circumvent this problem, synthetic data generation offers to generate a synthetic dataset that follows the same distribution as the underlying data while preserving the privacy of individuals.

A range of methods for synthetic data generation can be framed within the select-measure-update paradigm~\cite{McKenna_Miklau_Sheldon_2021, liu2021iterative}.  Such approaches iteratively \textit{select} a (linear statistical) query to answer, such as computing a particular $k$-way marginal distribution, \textit{measure} a noisy differentially-private evaluation of the selected query, and \textit{update} a probabilistic generative model using the noisy measurements.  One spends privacy budget to select and measure queries with substantial differences between model and data and repeatedly align the model with the noisy queries.

Different methods within this paradigm implement these steps in different ways.  A recent representative with good performance is AIM~\citep{mckenna2022aim} that models tables with discrete columns.  AIM is an advanced variant of approaches that placed first and second in the 2018 and 2020 NIST Differential Privacy Synthetic Data Challenges respectively.  AIM utilizes a clever selection procedure to identify improvable $k$-way marginals, measures these noisy marginals, and updates a probabilistic graphical model trained to match these marginals.  Marginals are convenient for privacy calculations, as each row in a tabular dataset appears in one cell of a marginal (so the sensitivity is typically $1$), and any computation that uses at most $k$ columns can be answered from $k$-way marginals.  Marginals are hence popular in the select-measure-update paradigm (see for example, ~\citet{tao2021benchmarking}).

Generally, $k$-way marginals with a low value of $k$ are selected due to privacy-accuracy tradeoffs and complexity reasons.  Low values of $k$ correspond to low-dimensional spaces on which it is easier to obtain good private statistics while higher values of $k$ correspond to more complex statistics that require higher privacy budgets and memory, as the number of cells in a marginal distribution scales up exponentially with $k$.  Additionally, for graphical models such as that used by AIM, the (exact) inference complexity scales exponentially with the junction tree width of selected marginals, and can only be limited by carefully selecting marginals with low $k$.

\begin{figure}[t]
\begin{center}
\includegraphics[width=0.5\textwidth]{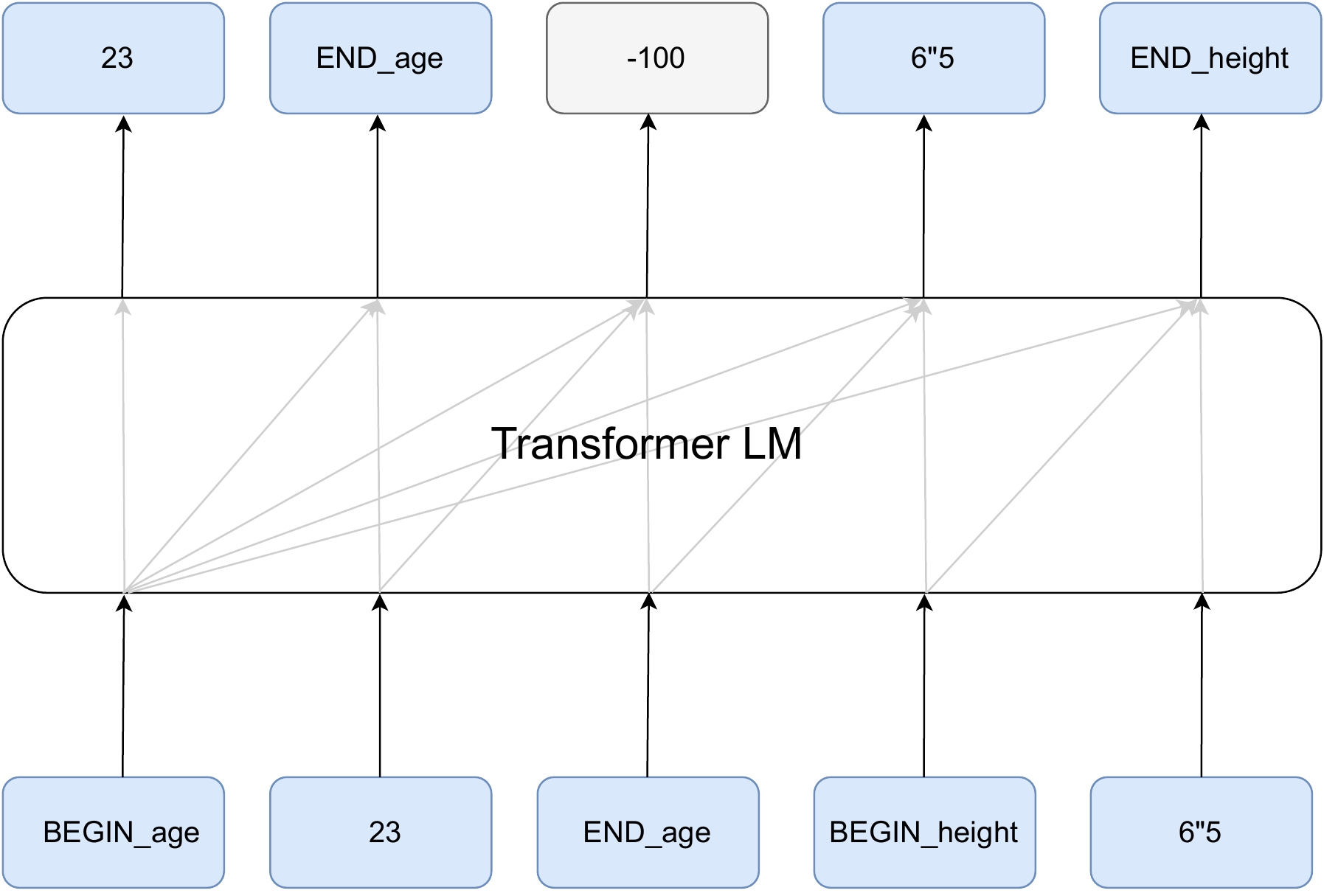}
\end{center}
\caption{\label{fig:lm} Example of transformer LM modelling a row where the column ``age" has value ``23" and the column ``height" has value ``6"5". 
SynLM is not trained to predict the ``BEGIN\_" part of a field, as the specific order of fields is decided by the prompter.}
\end{figure}

In this short paper, we depart from the select-measure-update paradigm and from measuring marginal distributions.  We instead train (privately) a deep probabilistic model (a transformer) \cite{vaswani_2017} using well-known tools from language modelling (LM).  We call this approach SynLM.  We treat each row in a table as a sentence, and train a transformer to predict elements in the row one-by-one.  Figure~\ref{fig:lm} is an example of such a prediction.  We observe SynLM is more scalable, fits the data better, and still approximates low-order marginal statistics well.

Selecting and measuring $k$-way marginals can become impractical as the number of columns increase or the number of unique discrete values per column increases.  A large table with complex columns has many $k$-way marginals, each of which may contain many cells.  In particular, numeric columns need to be discretized to coarse, low-precision values.  Unlike marginal-based approaches, SynLM's scaling is not dependent on exponentially-sized marginal distributions, but instead is linear in the number of columns and logarithmic in the number of possible discrete categories.

Our experimental results show SynLM outperforms AIM in terms of likelihood for almost all datasets, and the gap increases for datasets with higher likelihoods (see Table~\ref{tab:main_result}). 
In other words, SynLM becomes comparatively better for datasets with more columns or more categories per column.  We also compare SynLM and AIM in terms of marginals.  Similar to past benchmarking of synthetic data generation (\citet{tao2021benchmarking}), SynLM lags AIM in capturing low-dimensional marginals.  SynLM's weaker performance at low-dimensional marginals in conjunction with improved likelihood indicates complex dependencies are relevant in our experimental datasets.

In total, our contributions are the following,

\begin{itemize}
\item We show we can train a transformer from scratch on tabular datasets with Differential Privacy and obtain better likelihoods than AIM.
\item We show our method preserves marginals to some extent despite not being explicitly trained to do so.
\item We propose combining a trie with the language model to improve its modelling performance. 
\end{itemize}

\section{Background and related work}

\paragraph{Differential Privacy}~\citep{dwork2006calibrating,dwork2014algorithmic} is the standard notion of privacy for analytics and machine-learning purposes.
Algorithms satisfying differential privacy limit the impact that each individual sample has on their outcome.

\begin{definition}
A (randomized) algorithm $\calM$ is $(\varepsilon, \delta)$-differentially private if for datasets $S$ and $S'$ that differ by only one element, and for any subset $R \subset \mathrm{Im}(\calM)$, we have
\begin{equation}
\prob(\calM(S') \in R) \leq \exp(\varepsilon) \prob(\calM(S) \in R) + \delta
\end{equation}
\end{definition}

\paragraph{Graph-based synthetic data methods.} 
AIM~\citep{mckenna2022aim} and similar methods such as MWEM+PGM~\citep{mckenna2019graphical} model the data distribution by a graphical model.
AIM iteratively queries (privately) $k$-way marginals and updates the graphical model to match these marginals.  As there can be many probability distributions compatible with given (noisy) marginals, AIM fits the unique distribution with maximum entropy.

\subsection{Deep learning models}

\paragraph{Deep generative models} comprise several large families: GAN, VAE, Diffusion models, etc.  \citet{leduc2021composable} propose to encode data using codecs and conditionally predict different columns in a table.
Our approach bears some similarities, but we train our models using differential privacy, tokenize all fields and evaluate likelihood, which makes different models comparable.

\paragraph{Auto-regressive models} such as language models~\citep{bengio2000neural} factorize a joint distribution as
$ p(x_1, \dots, x_T) = \Pi_{i<T}~p(x_i~|~x_1, \dots, x_{i-1})$ to predict variables auto-regressively one by one.
Such a factorization is intuitive for naturally one-dimensional data such as language or speech.
This approach has also been extended to images~\citep{chen2020generative} by ordering pixels or tokens in a raster.
Language models are now routinely used to generate multi-modal data such as web-pages~\citep{aghajanyan2022cm3} or speech and audio~\citep{agostinelli2023musiclm, borsos2022audiolm}.

\paragraph{Training deep models with Differential Privacy.}
DP-SGD~\citep{abadi2016deep} is a modification of SGD (Stochastic Gradient Descent) that makes it differentially private.
It only requires to clip per-sample gradients and add Gaussian noise to the resulting sum.
To enjoy tight guarantees~\cite{mironov2019r}, samples in a batch are chosen uniformly at random (Poisson sampling).
Training with differential privacy implies a lower signal-to-noise ratio than a non-private equivalent.
This initially made non-deep alternatives look competitive~\citep{tramer2020differentially}, but recent work has shown that it is possible to obtain decent performance with deep models trained with DP.
In particular, large batch sizes~\citep{li2021large, anil2020dpbert} and data augmentations~\citep{de2022unlocking} are key ingredients to obtain good performance. 
To simulate efficiently large batch sizes for a large number of steps, \citet{sander2022tan} propose to run experiments at a lower batch size while keeping the same signal-to-noise ratio ($\sigma / q$).
\citet{sander2022tan} also show that it is worth doubling the batch size and the noise level $\sigma$ until $\sigma>2$, a fact also shown by other work in different forms~\citep{li2021large,tramer2020differentially}.
Training with large batches is one of the key ingredients behind the recent success of DP training of deep learning models~\citep{anil2020dpbert,de2022unlocking,sander2022tan}.

\begin{figure}[t]
    \centering
    \includegraphics[width=0.49\textwidth]{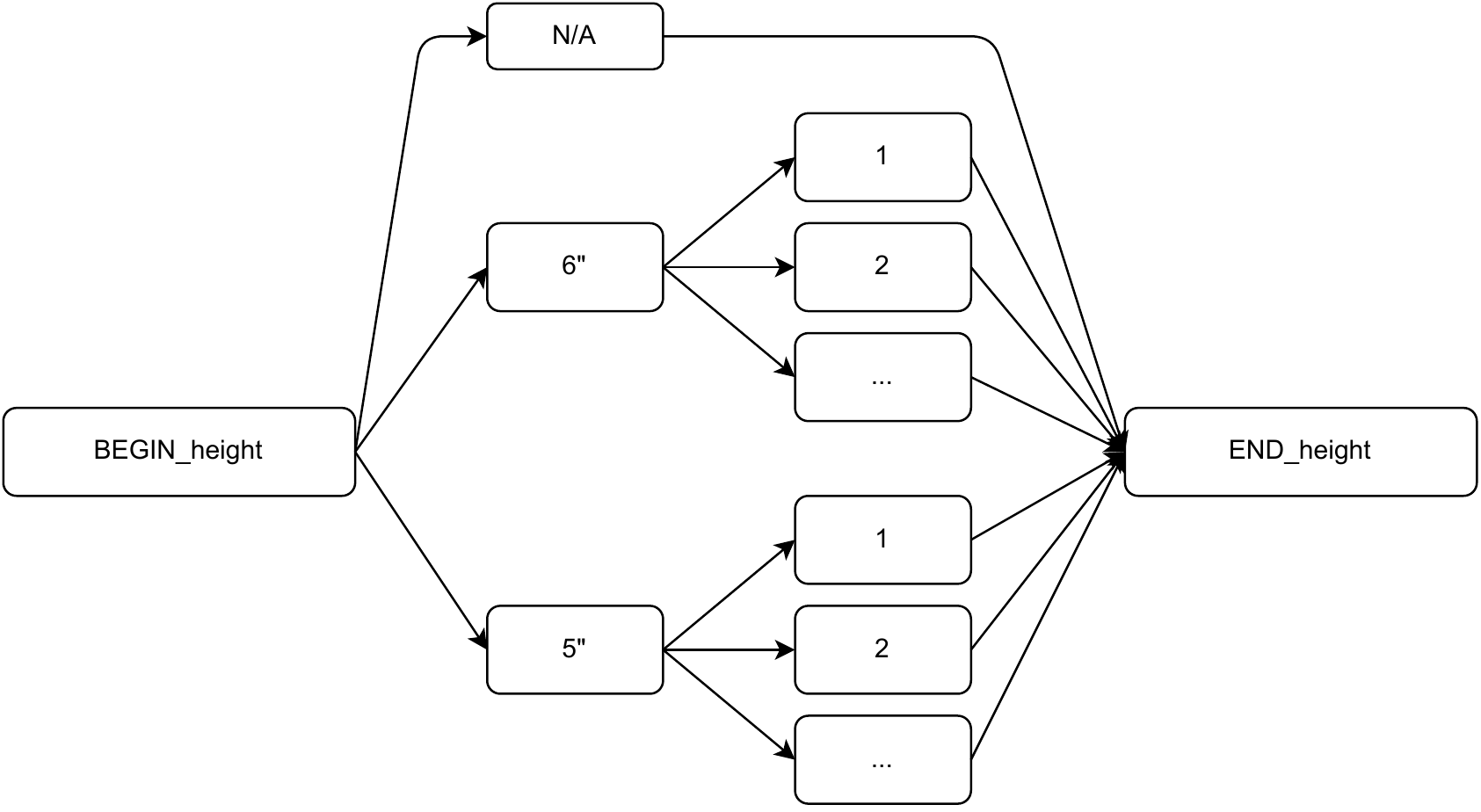}
    \caption{\label{fig:trie}
    Trie corresponding to a field ``height".
    SynLM only scores the possible value at each stage of the trie, instead of scoring the whole vocabulary as is customary for LMs.
    }
\end{figure}


\section{Our method}

\paragraph{Problem setting.}
Our data is given as a typical SQL table with $n$ rows and $C$ columns.  We assume each column takes value in a finite set $D_i$ (with $|D_i| \leq D$).
We consider that rows are i.i.d. samples from a distribution that can be entirely described by probability $\prob(c_1=d_1, c_2=d_2, \dots, c_C=d_C)$ for each $d_i \in D_i$.
Our objective is to estimate this distribution privately in order to generate synthetic samples, and we evaluate the quality of our estimation by computing the (mean) negative log-likelihood of held-out data from the same distribution.
We define $k$-way marginals as $\prob(c_{i_1}=d_{i_1}, c_{i_2}=d_{i_2}, \dots, c_{i_k}=d_{i_k})$ where $i_1, \dots, i_k$ are $k$ distinct integers in $[1, C]$.
For comparison with marginal-based approaches that primarily aim at matching queries on the  table, we also consider how well marginals on the original table are preserved in our generated data.
We consider DP guarantees with respect to adding or removing one sample, i.e. row from the table.

\subsection{Model}

Our transformer model SynLM is shown in Figure~\ref{fig:lm}.
It predicts fields one by one, and the order of the fields is decided when prompting the LM.
We use the GPT transformer model from the transformers library~\citep{wolf-etal-2020-transformers}.

\paragraph{Discretization.}
We discretize columns according to the following process to ensure that each column has at most cardinality $D=100$.
For columns of type string, we count the occurrence of each string and keep only the $99$ most frequent values, and bin the rest to an ``unknown" category. 
For integers and floats, we use SciPy's ``KBinsDiscretizer" with $100$ bins.
We emphasize that this part of the process is not differentially private and is not counted towards the overall budget. 
This is fine for our investigation, as we compare AIM and SynLM after this step of discretization which is the same for both algorithms.

\paragraph{Tokenization.}
We consider two schemes for tokenizing column values. 
Our first option is to use the ``level" tokenization, mapping each column value to an index from $0$ to $99$, and adding an offset so that values of different columns do not overlap.
The second option, ``semantic" tokenization, uses a (pre-trained) tokenizer such as BPE~\citep{sennrich2015neural} or SentencePiece~\citep{kudo2018sentencepiece} that converts a string $s$ to sequences of integers $\tok(s) = [t_1, \dots, t_p]$.

\paragraph{Converting a tokenized row into a sentence.} 
For each column, we add as prefix to its value a column prompt (in the form \texttt{BEGIN\_\{column\}}) and as suffix a column end token (\texttt{END\_\{column\}}). We then concatenate these columns either in the natural order of the columns or in a randomized order.  Note that during training we do not ask the model to predict the column prompt (\texttt{BEGIN\_\{column\}}) as it is fed to the model directly.

\paragraph{Prediction}
During training, the model predicts all tokens except \texttt{BEGIN\_} tokens, as this token is fed to the model as a prompt to predict the field.  We replace \texttt{BEGIN\_} tokens in the target by $-100$ tokens, which is a special token in PyTorch for which no loss is computed (and thus no gradient is backpropagated).

\paragraph{Augmentation}
During training, we augment our data and model and use AugMult~\citep{de2022unlocking}, aggregating augmented gradients before clipping (see Section~\ref{sec:training_dp}).
Data augmentation takes the form of randomly swapping the order of the fields (it is only active when we do not use a fixed field order).
Model augmentation corresponds to a different random seed for the random parts of the model like dropout.

\begin{figure}[t]
    \centering
    \includegraphics[width=0.49\textwidth]{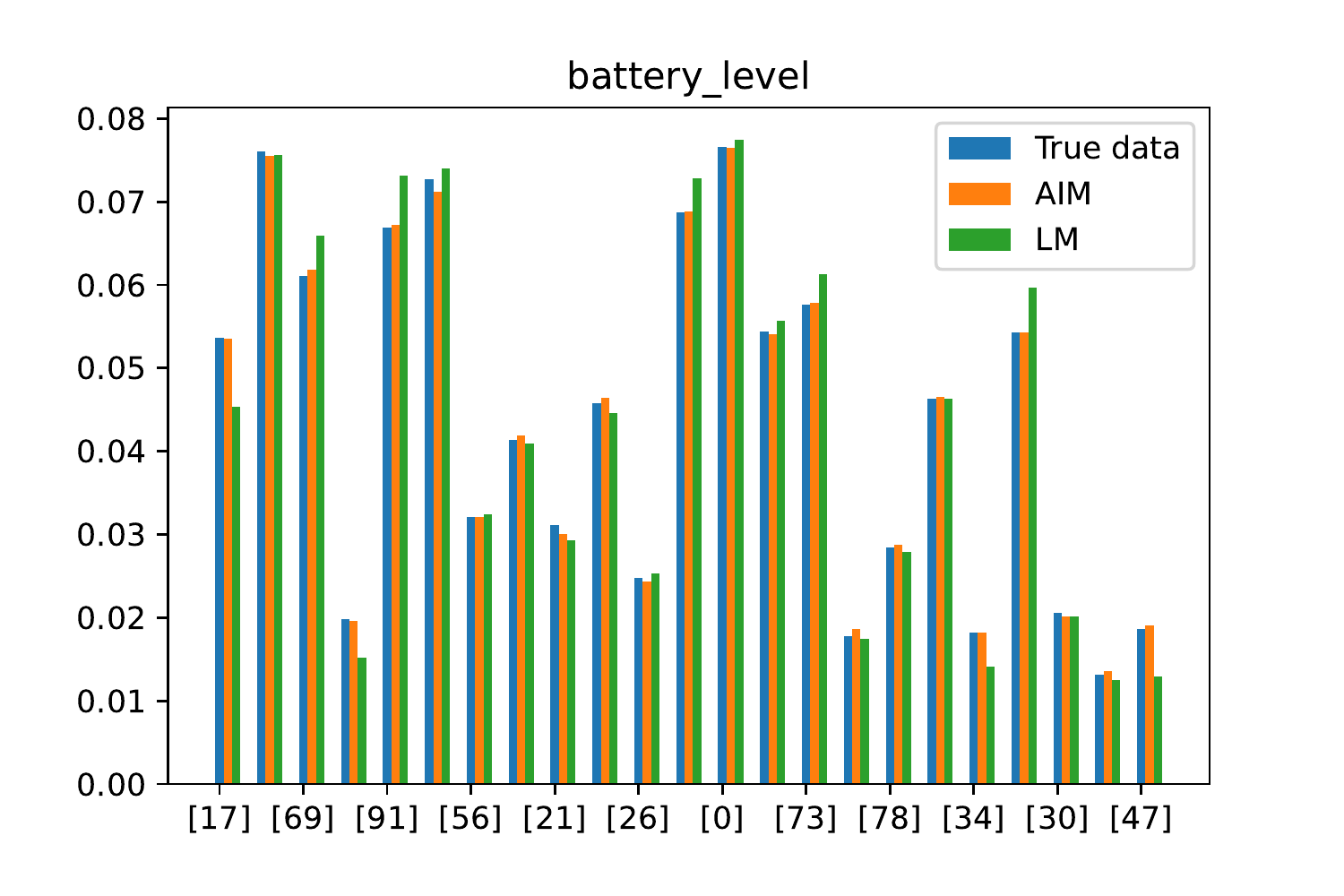}
    \caption{
    \label{fig:marginals}
    A representative example of 1-way marginals, computed for battery\_level on the scooter dataset.
    Marginals from the original data are shown in blue, from AIM in orange and from the LM in green.
    We can see that the LM obtains good approximate marginals, but AIM is much more precise.
    We follow previous research and compare the marginals on the synthetic data to the marginals of the full dataset (train + valid).
    }
\end{figure}

\subsection{Training with Differential Privacy}

Getting good performance while training a model with DP-SGD can be tricky.  We rely upon recent methods to get quality results with DP-SGD.

\label{sec:training_dp}
\paragraph{AugMult.}
\citet{de2022unlocking} propose a way to use augmentations in the context of DP-SGD.
Given a sample $z$, we consider (randomly) augmented models and the corresponding losses $l_1, \dots, l_j$ for each augmentation in the set $A(z)$ of augmentations of $z$. 
The average gradients is clipped to a constant $C$ ($\clip_C$) and noise is added with variance $\sigma^2 C^2$.
\begin{equation}
    \tilde{g} = \frac{1}{B} \left( \sum_{i=1}^B \clip_C \left( \frac{1}{p}\sum_{j \in \Aug(z_i)} \nabla_\theta \ell_j \right) + \gauss(0, \sigma^2 C^2) \right) \nonumber
\end{equation}

\paragraph{TAN.}
\citet{sander2022tan} propose a method to choose hyper-parameters for DP training using low-compute simulations.
Given a target batch size $B$, and noise level $\sigma$, they propose to scale down the batch size to $B' \ll B$ and noise $\sigma' = \sigma B'/B$.
This approach allows to cross-validate hyper-parameters while limiting the compute expenditure. 
In this work, we choose the same set of hyper-parameters on all datasets and can thus choose these on smaller datasets.

\subsection{Constrained generation using a trie}
Auto-regressive models are prone to generating invalid values: for every prediction, there is a small chance that the LM predicts an invalid token (say, predict $1000$ for age), and that probability compounds over the length of the generated sequence.
We prevent this by forcing the model to only output valid (sequences of) tokens. 
We build a trie~\citep{fredkin1960trie, de1959file}  using all possible values for each column. 
We then guide the LM by only allowing it to sample among valid nodes in the trie. 
The trie structure is naturally compatible with the left-to-right factorization of language models: at each timestep, the trie restrains the list of possible outputs and we restrict the output of the language model to be in this list of possible values by setting the probability of other values to $0$.  An example trie is shown in Figure~\ref{fig:trie}.

We use this trie at inference time (generation and evaluation), and also for training.  At inference, it guarantees the model outputs a valid sequence (generation) or that the probability of a sequence is normalized only among valid sequences (evaluation).  For training, as it forces the model to only compare the correct prediction to other valid predictions, it ensures the model focuses on useful signal.  We note that the trie reduces to a simple structure for level tokenization as each field is mapped to exactly one token.

\begin{table}[t]
\centering
\begin{tabular}{lrr}
\toprule
Dataset & Rows & Columns \\
\midrule
adult & 32,561 & 15 \\
car & 1,728 & 7 \\
covtype & 581,012 & 55 \\
scooter & 27,113 & 6 \\
intrusion & 494,021 & 41 \\
shopping & 12,330 & 18 \\
marketing & 41,188 & 21 \\
\bottomrule
\end{tabular}
\caption{Size of the datasets.}
\label{tab:dataset}
\end{table}
\subsection{Scaling properties}
\label{sec:scaling}

We describe (roughly) how both \synlm and marginal-based methods scale with the number of columns $C$ or the number of unique values per column $D$.  Marginal-based methods need to compute all $k$-way marginals: there are $\binom{C}{k}=\bigO(C^k)$ such marginals, and each marginal is a histogram over $D^k$ distinct values. 
This leads to a complexity (time and space) of $\bigO(C^k D^k)$.
In practice, this complexity is capped by the number of unique values in the table of which a trivial upper-bound is the number of rows $N$.
The overall complexity of just generating the data for a marginal-based method is thus at least $\bigO(C^k \min(D^k, N))$.

On the contrary, \synlm scales linearly in the number of columns $C$, and scales as $\log(D)$ in the number of categories.
Indeed, the language model uses a fixed tokenization, let us call $V$ its vocabulary size. 
The average length of a tokenized field with $D$ distinct values is $\frac{\log(D)}{V}$.
The number of operations required to train a transformer model is directly proportional to the number of tokens seen~\citep{kaplan2020scaling}.
Given that each of the $N$ rows is seen for $E$ epochs and has averaged tokenized length $\frac{C \log(D)}{V}$, training the model thus requires $\bigO \left(N\times E \times C \log(D) \right)$ operations.


\section{Experiments}

\begin{table}[t]
    \centering
\begin{tabular}{ccc}
\toprule
Dataset & AIM & SynLM \\
\midrule
intrusion & 5.8 & \textbf{4.7} \\
scooter & \textbf{5.8} & \textbf{5.8}\\
car & \textbf{7.8} & 8.0 \\
adult & 18.9 & \textbf{18.6}\\
marketing & 19.0 & \textbf{18.8}\\
shopping & 28.0 & \textbf{22.9}\\
covtype & 36.2 & \textbf{28.5}\\
\bottomrule
\end{tabular}
\caption{\label{tab:main_result}
Comparison of our method (SynLM) and the state-of-the-art AIM.
We report negative log-likelihood on a held-out set, where smaller is better and shown in bold.  \synlm is better than AIM on almost all datasets (except scooter and car). 
}
\end{table}

\begin{table*}[t]
    \centering
    \begin{tabular}{cccccc}
    \toprule
    Dataset & \synlm & Random order & No trie guiding & Semantic tokenization & GPT-2 \\
    \midrule
    scooter & \textbf{5.8} & 5.8 & 7.7 & 5.8 & 5.9\\
    car & \textbf{8.0} & 8.4 & 8.5 & 8.1 & 8.2\\
    adult & \textbf{18.6} & 19.1 & 20.1 & 18.7 & 18.8 \\
    marketing & \textbf{18.8} & 19.8 & 20.9 & 18.9 & 19.2 \\
    shopping & \textbf{22.9} & 24.3 & 34.4 & 23.6 & 24.1 \\
    \bottomrule
    \end{tabular}
    \caption{\label{tab:ablation}
    Comparison of \synlm using a fixed order of columns, trie guiding, and ``level" tokenization, with alternatives including random ordering, removing the trie, semantic tokenization, and larger GPT-2 sized models.  We report negative log-likelihood on a held-out set, where smaller is better and is shown in bold.
    }
\end{table*}

\paragraph{Datasets.}
We evaluate our method on multiple datasets. 
We use the datasets Adult, Car, Covertype, Marketing and Shopping from the UCI machine learning repository~\citep{dua2019uci}, the Scooter dataset from \citet{scooterdataset} and the Intrusion~\citep{intrusiondataset} dataset from \citet{sdgym}.
For each dataset, we carve-out as validation set $10\%$ of the samples or $10,000$ samples, whichever is smaller.
Table~\ref{tab:dataset} references the sizes of each dataset: the number of rows varies from $1,728$ to $581,012$ and the number of columns from $6$ to $55$.

\paragraph{Language model.}
We train a GPT transformer using the transformers~\citep{wolf-etal-2020-transformers} library, with Pytorch~\citep{paszke2019pytorch} and Opacus~\citep{yousefpour2021opacus}.
We train a transformer with $4$ layers, $d=256$, $4$ heads for $100,000$ steps using AdamW and a batch size $B$ chosen automatically on each dataset to ensure $\sigma>2$.
We use a linear learning rate schedule going down from $5 * 10^{-4}$ to $0$ with a warm-up of $100$ steps.
We set the noise level $\sigma$ such that $\varepsilon=5$ for $\delta=10^{-6}$, which are common choices for privacy parameters ($\delta<1/n$ and $\varepsilon \in [1, 10]$).
We chose the dataset ``car" to tune hyper-parameters (learning rate, number of steps, warm-up).
The optimal hyper-parameters themselves could theoretically reveal information about the dataset.
To limit such leakage, we chose one dataset to validate the hyper-parameters and in a practical setting we would tune hyper-parameters on a public set and just perform one private training on private datasets.

\paragraph{AIM}
We use AIM with standard parameters, using the same privacy budget as for \synlm ($\varepsilon=5, \delta=10^{-6}$).
To increase fairness with the tuned hyper-parameters of SynLM\footnote{We also explored increasing the maximum number of cells and maximum memory constraints from AIM defaults of 100k and 80 MB to 1M and 5 GB respectively, and results were the same or worse except on Scooter which was slightly improved.}, we pick the $k$-way marginal parameter with the best results, selecting from $k \in \{1, 2, 3\}$ for each dataset.

\subsection{Results}

We report results on \synlm trained with a fixed order of columns and the ``level" tokenization.
The overall results show that \synlm is better on most datasets in terms of likelihood (see Table~\ref{tab:main_result}), but AIM is substantially better at computing marginals.

\paragraph{Likelihood evaluation}
Table~\ref{tab:main_result} shows the negative log-likelihood given by AIM and \synlm on our benchmark.
We can see that \synlm is better on all datasets except one, and that the difference is small for most datasets except shopping and covtype.
It seems that the LM is comparatively better for datasets with higher log-likelihoods (i.e. more complex data), we hypothesize that this is probably linked to the scaling properties discussed in Section~\ref{sec:scaling}.

\paragraph{Marginal evaluation}
We observe synthetic data generated using \synlm present empirical marginals that are qualitatively close to marginals of the original data.
Figure~\ref{fig:marginals} shows an example of $1$-way marginals computed on the scooter dataset, and the ones computed on the generated data.
This finding is consistent with \citet{canale2022generative}, who considered mostly non-private language models.

\subsection{Ablations}

We consider several variants of our method and analyze them on all datasets (except intrusion and covtype as these are the largest datasets), and report the results in Table~\ref{tab:ablation}.

\paragraph{Ordering} fields in a fixed order seems to be better than a random order (different every time).
We hypothesize that a fixed order allows the model to focus on precise dependencies, whereas a randomized order would require the LM to model arbitrary dependencies, making its task more complex.
It is also possible that the default order of columns corresponds to a natural way of modelling the data and thus performs better.

\paragraph{Tokenization scheme.}
Semantic tokenization seems worse than level tokenization, although by a smaller margin.
We conjecture that this is because level tokenization differentiates tokens for each column.
For example, if two columns are numerical, the semantic tokens corresponding to a fixed value (say "40") will be the same although they can have different meaning (for example, 40 dollars is different from age 40).
On the contrary, level tokenization will assign a different token to "40" for each column.

\paragraph{Trie guiding} is essential to our method. 
Indeed, models trained and evaluated without trie guiding have significantly worse likelihoods.
We also found that training without the trie and evaluating with the trie is worse than doing both with the trie but we did not report the results for brevity.

\paragraph{Using GPT2-size models.}
Finally, we tested larger models (GPT2-size with 131M parameters).
We find that these models, for which we used semantic tokenization, have an overall worse likelihood than smaller models.
The datasets may be too small to warrant using such large models trained from scratch.

\section{Conclusion}

In this short paper, we proposed an approach for privately generating synthetic data from tabular data using language models.  Our approach, SynLM, trains a transformer language model on the tabular data and generates synthetic data that preserves the privacy of individuals.  We showed SynLM outperforms a state-of-the-art marginal-based method in terms of likelihood, is more scalable, and still approximates low-order marginal statistics somewhat.
Overall, this initial research indicates that private language modeling offers a promising direction for generating synthetic tabular data.

We specifically point out some limitations of our work and opportunities for future research.  First, training state-of-the-art models with DP-SGD requires large batches~\citep{li2021large,sander2022tan}.  This is not a limitation of the present study since the experimental datasets are small, but could limit applying SynLM directly on larger datasets.  Second, a broader range of comparisons is certainly possible; our comparison to AIM itself is not exhaustive and other ablations, for example lower privacy budgets or different discretization, could be performed.  Private language modeling could also be compared against more methods than just AIM in the future.  Third, our focus on held-out likelihood seems untraditional within the synthetic data generation literature, which often focuses on producing synthetic datasets that match results of computations, often answerable by low-order marginals, on the private dataset.  Comparing how other past synthetic data methods perform on held-out likelihood could also be useful.  Finally, we only consider training language models from scratch with a particular sentence prompt and do not leverage pre-trained language models.  We conjecture that a semantic prompting of a table row as a sentence combined with a pre-trained (large) language model and private fine-tuning could improve the performance of private language modeling for tabular data even further.

\section*{Acknowledgements}
The authors thank Ryan McKenna for helpful discussions, and in particular his code to compute likelihoods (see Appendix~\ref{code:likelihood}).

\bibliography{main}
\bibliographystyle{icml2022/icml2022}

\newpage
\onecolumn
\appendix
\section{Computing likelihood}
\label{code:likelihood}
\begin{lstlisting}[language=Python, caption={Code to compute the likelihood of data under a PGM model.}]
def compute_negative_log_likelihood(model, data):
    logZ = model.belief_propagation(model.potentials, logZ=True)
    log_probas = np.zeros(data.records)

    for cl in model.cliques:
        P = model.potentials[cl].values
        idxs = data.project(cl).df.values.astype(int)
        log_probas += np.array([P[tuple(i)] for i in idxs])

    return logZ - log_probas
\end{lstlisting}

\end{document}